\documentclass[11pt,a4paper]{article}
\usepackage[hyperref]{emnlp2018}
\usepackage{mathptmx}
\DeclareMathAlphabet{\mathcal}{OMS}{cmsy}{m}{n}
\usepackage{latexsym}
\usepackage{mathtools}
\usepackage{booktabs}
\usepackage{multirow}
\usepackage{multicol}
\usepackage{url}
\usepackage{amssymb}
\usepackage{dsfont}
\usepackage{pgfplots}
\usepackage{pgfplotstable}
\usepackage{footmisc}
\usepackage{stfloats}
\usepackage{placeins}
\usepackage{caption}

\usepackage{tikz}
\usetikzlibrary{shapes}
\usetikzlibrary{decorations.pathreplacing}
\usetikzlibrary{arrows.meta}
\tikzset{operator/.style={circle, draw, inner sep=0pt, minimum size=.8cm}}
\tikzset{font={\fontsize{9pt}{12}\selectfont}}
\definecolor{PaleBlue}{rgb}{0..55,.9}
\definecolor{PaleGreen}{rgb}{0..7,.25}
\definecolor{RedPink}{rgb}{.9,0..2}
\definecolor{Pink}{rgb}{.8,.2,.6}
\definecolor{Purple}{rgb}{.6,0..75}
\definecolor{Orange}{rgb}{.9,.3,.05}

\tikzstyle{embed}=[%
    draw,
	#1,
	thick,
	anchor=north,
	minimum width=.3cm,
	minimum height=.6cm,
	inner sep=0pt,
	text=#1!65!black
	]

\definecolor{color_conll}{rgb}{.1,.1,.8}
\definecolor{color_semeval}{rgb}{.1,.8,1}
\definecolor{color_glove}{rgb}{.75,.15,.65}

\tikzstyle{comick_style}=[opacity=.6,line width=1pt, solid, #1, mark size=1.2pt, mark=*, mark options={#1, solid}]
\tikzstyle{mimick_style}=[opacity=.6,line width=1pt, #1, dashed, dash pattern=on 7pt off 1.5pt on 1pt off 1.5pt, mark size=1.2pt, mark=*, mark options={#1, solid}]

\let\primee=\prime
\renewcommand{\prime}{\hspace{1pt}^\primee}

\aclfinalcopy 

\title{Predicting and interpreting embeddings for out of vocabulary words in downstream tasks}

\author{Nicolas Garneau\thanks{\hspace{4pt}Authors contributed equally to this work.},\hspace{5pt}Jean-Samuel Leboeuf\footnotemark[1],\hspace{5pt} Luc Lamontagne\\
 D\'epartement d'informatique et de g\'enie logiciel\\
 Universit\'e Laval, Qu\'ebec, Canada\\
 \texttt{\{nicolas.garneau, jean-samuel.leboeuf\}.1@ulaval.ca},\\ \texttt{luc.lamongtagne@ift.ulaval.ca}
 }

\date{}

\begin{document}
\maketitle
\begin{abstract}

We propose a novel way to handle out of vocabulary (OOV) words in downstream natural language processing (NLP) tasks.
We implement a network that predicts useful embeddings for OOV words based on their morphology and on the context in which they appear.
Our model also incorporates an attention mechanism indicating the focus allocated to the left context words, the right context words or the word's characters, hence making the prediction more interpretable.
The model is a ``drop-in'' module that is jointly trained with the downstream task's neural network, thus producing embeddings specialized for the task at hand.
When the task is mostly syntactical, we observe that our model aims most of its attention on surface form characters.
On the other hand, for tasks more semantical, the network allocates more attention to the surrounding words. 
In all our tests, the module helps the network to achieve better performances in comparison to the use of simple random embeddings.


\end{abstract}

\section{Introduction and motivation}



\citet{goldberg2017neural_chap8} emphasizes the fact that out of vocabulary (OOV) words represent a problem often underestimated for NLP tasks such as part of speech tagging (POS) or named entity recognition (NER) \cite{collobert2011natural, turian2010word}.
Due to the lack of proper ways to handle OOV words, researchers often resort to simply assign random embeddings to unknown words or to map them to a unique ``unknown'' embedding, hoping their model will generalize well nonetheless.


An interesting way to handle OOV words is the Mimick model \cite{pinter2017mimicking}.
This model aims to predict embeddings such as GloVe \cite{pennington2014glove} for OOV words by training a recurrent network on the characters of the words.
While being simple, this model improves the accuracy of POS tagging as well as morphosyntactic attribute tagging on the Universal Dependencies corpus \cite{de2014universal}.

We propose an extension to this model by taking into account not only the surface form of a word (i.e. its characters) but also the embeddings of its surrounding words.
We hypothesize that context words provide useful semantic and syntactic information to model unknown word embeddings, hence complementing cues given by its characters. 
For this purpose, we introduce a module that can make, for the same word in different contexts, different predictions. It can also learn ``specialized'' embeddings for a specific downstream task which we evaluate for two sequence labeling tasks.
Furthermore, we add to our model an attention/interpretation mechanism to determine which of the left context, right context or the surface form of a word receives more attention during prediction. Our experimental results are depicted in a quantitative and qualitative analysis.



\section{Architecture}

To test our ideas, we developed an OOV prediction module comprising the following components. First, the left context, right context and word characters are fed to three bi-LSTMs to produce separate encodings. These three hidden states are then passed to a linear layer on which a softmax is applied to determine their relative importance (i.e. their degree of attention). The output of this layer is then used to produce a weighted sum of the hidden states. Finally, a simple layer computes an embedding from this sum.

To evaluate the contribution of this OOV prediction scheme to sequence labeling tasks, we use a bi-LSTM architecture on the resulting word embeddings and apply a softmax on the hidden state of each word to predict tags.

\begin{table}[t!]
\begin{center}
\begin{tabular}{c c c c}
\toprule
Task & Metric & Random Emb. & Our module\\
\midrule
NER      & F1   & 77.56 & \textbf{80.62} \\
POS      & acc. & 91.41 & \textbf{92.58} \\
\bottomrule
\end{tabular}
\end{center}
\renewcommand\thetable{1}
\caption{%
Comparison of our OOV embeddings prediction scheme against random embeddings.
}\label{tab:ext_baseline_metrics_minimal}
\end{table}

\begin{table}[t!]
\begin{center}
\setlength{\tabcolsep}{5.5pt}
\begin{tabular}{c l c c c c}
\toprule
Task & Tag & Ex & Word & Left & Right\\
\midrule
\multirow{9}{*}{NER} & O           & 1039  & \textbf{0.81}    & 0.08    & 0.11 \\
& B-PERS      & 63    & 0.21    & 0.31    & \textbf{0.49} \\
& I-PER	    & 119   & 0.16	& \textbf{0.52}	& 0.32 \\
& B-ORG	    & 40	& 0.26	& 0.30	& \textbf{0.44} \\ 
& I-ORG	    & 3	    & 0.27	& 0.31	    & \textbf{0.42} \\
& B-LOC	    & 13	& 0.23	    & 0.30	& \textbf{0.47} \\
& I-LOC	    & 2	    & 0.16	& \textbf{0.48}	& 0.36 \\
& B-MISC	    & 47	& \textbf{0.40}	& 0.21	& 0.39 \\
& I-MISC	    & 5	    & \textbf{0.41}	& 0.26	& 0.33 \\
\midrule
\multirow{5}{*}{POS} & NNP	& 308	& 0.29	& 0.31	& \textbf{0.40} \\
& NN	& 46	& \textbf{0.45}	& 0.20	& 0.35 \\
& CD	& 827	& \textbf{0.86}	& 0.05	& 0.09 \\
& NNS	& 23	& \textbf{0.37}	& 0.24	& \textbf{0.39} \\
& JJ	& 100	& \textbf{0.49}	& 0.15	& 0.36 \\
\bottomrule
\end{tabular}
\end{center}
\renewcommand\thetable{2}
\caption{%
Average weights assigned to word characters, left context and right context by the attention mechanism for NER and for POS tagging.
}\label{tab:attn_ner}
\end{table}

\section{Experimental results and discussion}

We evaluate the performance gain that our module can offer by solving two sequence labeling tasks, NER and POS tagging, using the \textit{CoNLL 2003 shared task} dataset.
We compare our module to a baseline where OOV words are assigned random embeddings.
Table~\ref{tab:ext_baseline_metrics_minimal} shows the results we obtain.
We can observe the clear advantage of proper handling of OOV words can provide.
For both tasks, we gain a significant margin on the baseline, with more than 3\% of the F1 score for NER.

\begin{table*}[b!]
\begin{center}
\setlength{\tabcolsep}{5pt}
\begin{tabular}{l c c c c c c}
\toprule
Word & Left & Right & Examples \\
\midrule
0.24 &	\textbf{0.38} &	\textbf{0.38} &\small	\textbf{\texttt{<BOS>}} \textit{langmore} \textbf{, a persistent campaigner for interventionist economic} \\
0.15 &	\textbf{0.59} &	0.26 &\small	 \textbf{\texttt{<BOS>} australian parliamentarian john} \textit{langmore} has formally resigned from his lower house\\
0.15 &	\textbf{0.61} &	0.24 &\small	\textbf{had received today from mr john vance} \textit{langmore} , a letter resigning his place as \\
0.15 &	\textbf{0.69} &	0.16 &\small 	\textbf{\texttt{<BOS>} rtrs - australian mp john} \textit{langmore} formally resigns . \texttt{<EOS>}\\
0.17 &	\textbf{0.40} &	\textbf{0.43} &\small 	\textbf{\texttt{<BOS>}} \textit{langmore} \textbf{, 57 , announced in november that}\\
\bottomrule
\end{tabular}
\end{center}
\renewcommand\thetable{3}
\caption{%
Qualitative example on the OOV word \textit{langmore} which is an entity of type \texttt{PER}. We can cleary see that depending on the context, the weights may shift drastically.
}\label{tab:qualitative}
\end{table*}

We can see from Table~\ref{tab:attn_ner} that the network focuses more on the context for a semantic task such as NER.
An interesting phenomenon is a focus on the right context when the entity is of type \textit{B} and on the left context when the entity is of type \textit{I}.
We can also note that for the syntactic task (POS), the network tends to focus on the context for proper nouns (NNP), which corroborates our observations for the NER task.
However, morphology plays a more important role to predict embeddings for other lexical categories.
Embeddings for quantities (CD) are mostly predicted from their numerical characters.

We further qualitatively analyze the behavior of the network for a given OOV word appearing in different contexts in Table~\ref{tab:qualitative}.
When the target OOV word \textit{langmore} is preceded by \textit{john} or \textit{australian}, the network gives high importance to these context words.
However, an interesting phenomenon happens when a sentence begins with this word: the network shifts its attention from the left context to the right one and also assigns more importance to the morphology of the word, thus showing the network has truly learned where it can extract useful information.

\section{Future works}
In our future works, we plan to apply the attention mechanism specifically on the characters of the OOV word and the words that compose the context instead of using the hidden state of the respective elements only.
We are also looking forward to testing our attention model in different languages and on other NLP tasks such as machine translation.
We hope to present the full results and the architecture of our model in more details in a paper to be published relatively soon.

\clearpage

\bibliography{emnlp2018}
\bibliographystyle{acl_natbib_nourl}

\appendix

\end{document}